\documentclass[11pt]{article}
\pdfoutput=1
\usepackage{acl2015}
\usepackage{times}
\usepackage{url}
\usepackage{latexsym}
\usepackage{graphicx}
\usepackage{booktabs}
\usepackage{amsmath}
\usepackage{subfig}
\usepackage{textcomp}
\usepackage[bottom]{footmisc}

\title{Online Representation Learning in Recurrent Neural Language Models}

\author{Marek Rei \\
The ALTA Institute\\
Computer Laboratory\\
University of Cambridge\\
United Kingdom\\
{\tt marek.rei@cl.cam.ac.uk} \\}

\date{}

\begin{document}
\maketitle
\begin{abstract}
We investigate an extension of continuous online learning in recurrent neural network language models. The model keeps a separate vector representation of the current unit of text being processed and adaptively adjusts it after each prediction. The initial experiments give promising results, indicating that the method is able to increase language modelling accuracy, while also decreasing the parameters needed to store the model along with the computation required at each step.
\end{abstract}

\section{Introduction}

In recent years, neural network models have shown impressive performance on many natural language processing tasks, such as speech recognition \cite{Chorowski2014,Graves2013a}, machine translation \cite{Kalchbrenner2013,Cho2014a}, text classification \cite{Le2014a,Kalchbrenner2014} and image description generation \cite{Kiros2014}. One of the main advantages of these methods is the ability to learn smooth vector representations for words, thereby reducing the sparsity problem inherent in any natural language dataset. 

Language modelling is another task where neural networks have delivered excellent results \cite{Bengio2003,Kombrinka}.
\newcite{Chelba2014} have recently benchmarked several well-known language models by training on very large datasets. They found that a recurrent neural network language model (RNNLM) combined with a 9-gram MaxEnt model was able to give the best results and lowest perplexity.

In this work we investigate a possible extension of RNNLM, by allowing it to continue learning and adapting during testing. The model keeps a vector representation of the current sentence that is being processed, and continuously modifies it based on an error signal. We refer to this as a version of online learning, as gradient descent is used to optimise the vector even during testing. 

The technique is inspired by work on representation learning \cite{Weston2008,Mnih,Mikolov2013a}, especially \newcite{Le2014a} who use a related model to learn representations for text classification. We extend the idea to recurrent models and apply it to the task of language modelling. Our results indicate that by exchanging some existing model parameters for a component using online learning, the system is able to achieve lower perplexity while also reducing the necessary computation.

\section{RNNLM}

\begin{figure}[b]
	\centering
    \includegraphics[width=0.6\linewidth]{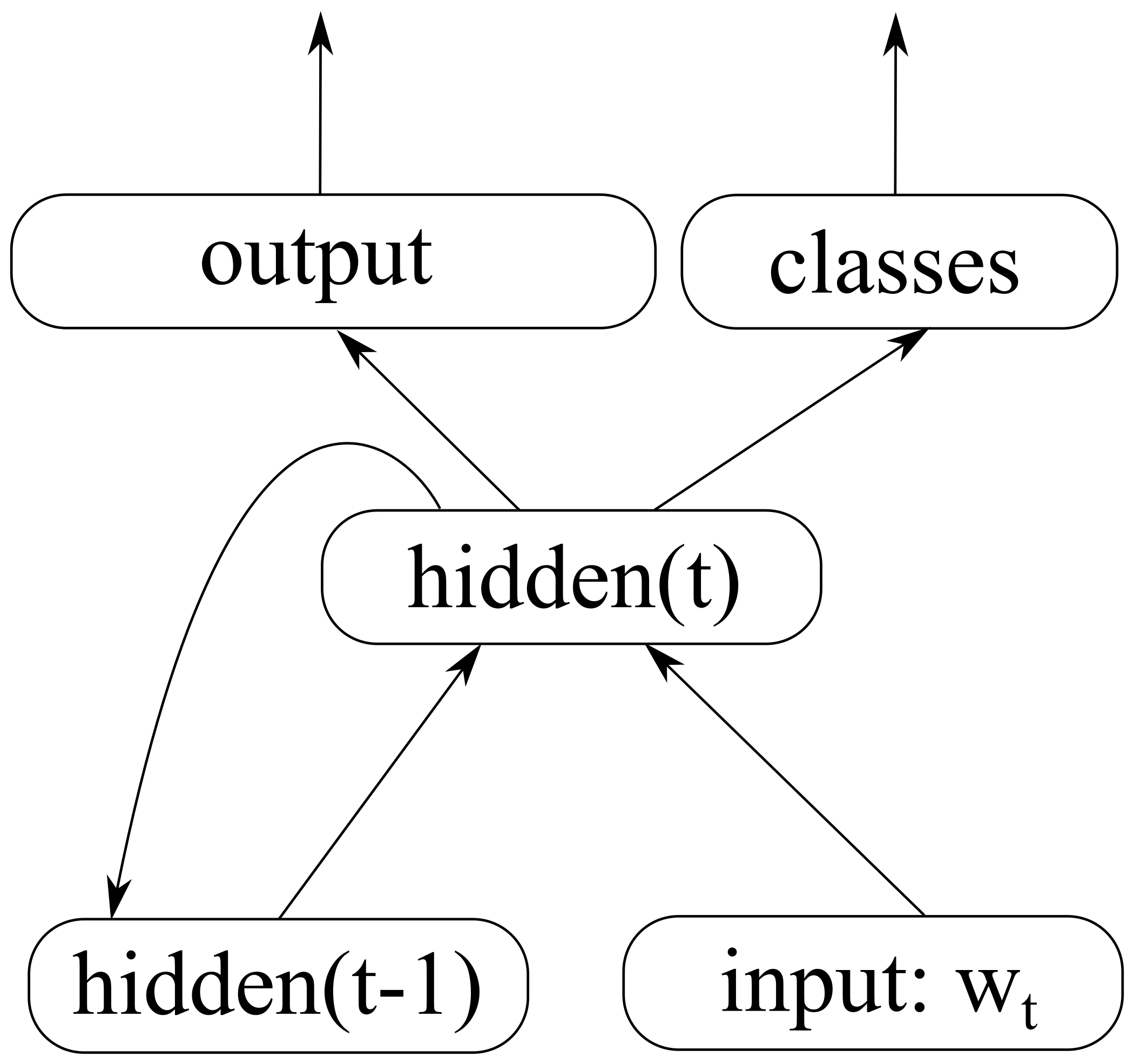}
	\caption{Recurrent neural network language model (RNNLM)}
	\label{fig:rnnlm}
\end{figure}

We base our implementation of the RNNLM on \newcite{Kombrinka}, shown in Figure \ref{fig:rnnlm}. The input layer to the network consists of a 1-hot vector representing the previous word in the sequence, and the hidden vector from the previous time step. These are multiplied by corresponding weight matrices and the resulting vectors are passed through an activation function to calculate the hidden vector at the current time step.\footnote{Explicit multiplication for the word vectors can be avoided by using data structures that retrieve the correct vector in constant time.}

Class-based output architecture is used to avoid calculating the softmax over all words in the vocabulary. 
The probability distributions over words and classes are calculated by multiplying the hidden vector with the corresponding weight matrix and applying the softmax function:
$$
hidden_t = \sigma(E \cdot input_t + W_h \cdot hidden_{t-1})
$$
$$
classes = softmax(W_c \cdot hidden_t)
$$
\vspace{-3.5mm}
$$
output = softmax(W_o^{(c)} \cdot hidden_t)
$$

\noindent where $\sigma$ is the logistic function and $W_o^{(c)}$ is the weight matrix between the hidden layer and the output words in class $c$.

Finally, we multiply the probability of the next word belonging to class $c$ with the output probability of the next word given the class to get the overall probability of the next word given the previous words:
$$
P(w_{t+1} | w_{1}^{t}) \approx classes_c \cdot output_{w_{t+1}}
$$

Negative log-probability is used as the loss function, which optimises the network to assign a high probability to the correct words.
The network is trained using gradient descent and backpropagation through time. In the basic model, this means unrolling the recurrent network for a fixed number of time steps, essentially turning it into a deep feedforward network which outputs probability distributions on different layers. Instead of using a fixed number of steps, our implementation unrolls each sentence from the last word to the first word, making it more suitable for processing individual sentences as opposed to longer texts. 

In addition, we introduce a special vector to use as the hidden vector at the start of each sentence.
The values in this vector are treated as parameters and optimised during training. This allows the network to learn a suitable starting point when no other information is available, giving slight performance improvements in our experiments.

\section{RNNLM with online learning}

\begin{figure}[t]
	\centering
    \includegraphics[width=0.6\linewidth]{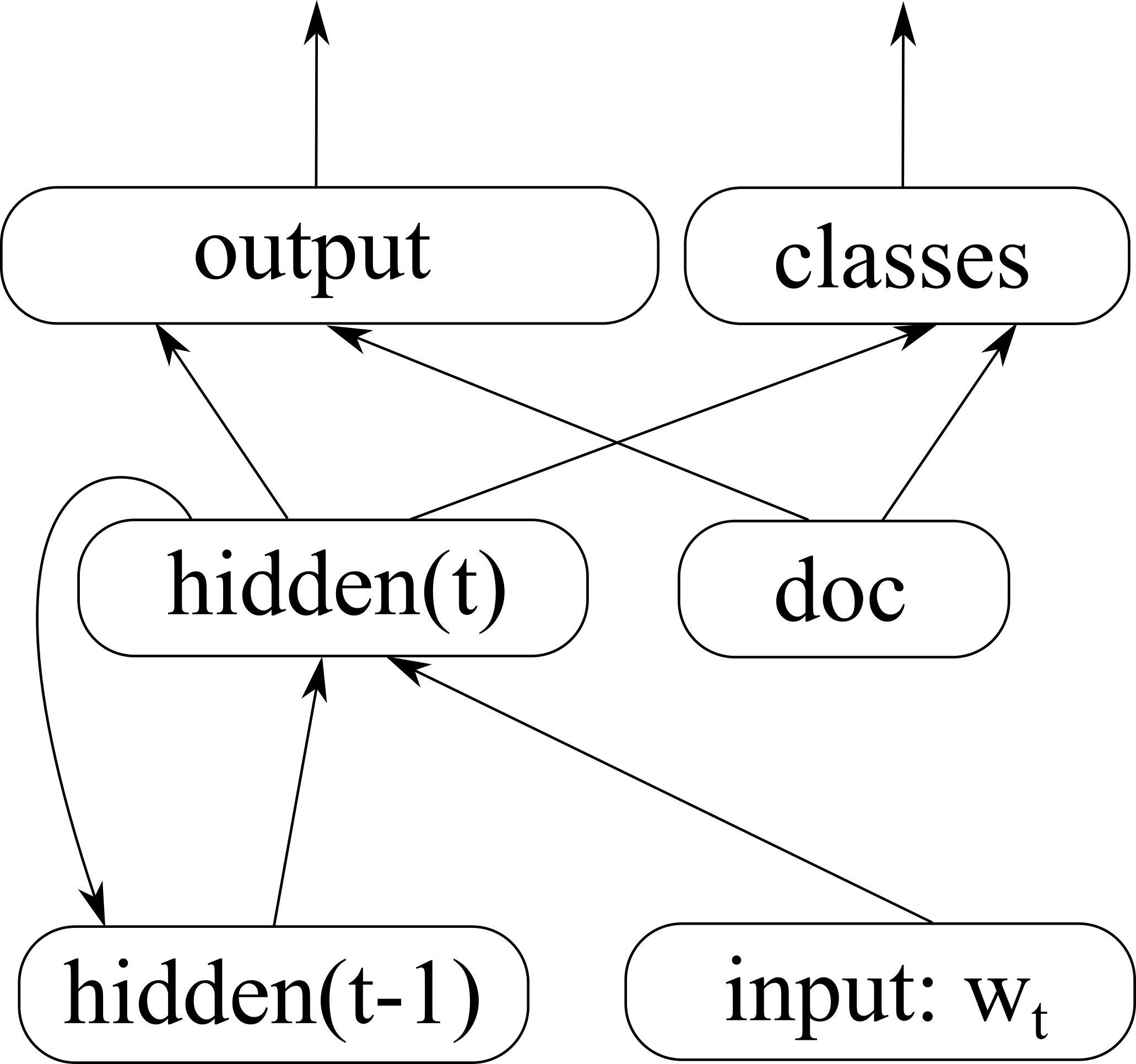}
	\caption{RNNLM with an additional document vector for active learning}
	\label{fig:bprnnlm}
\end{figure}

We extend the RNNLM by introducing an additional document/context vector, shown as $doc$ in Figure \ref{fig:bprnnlm}. This vector will represent the current document being processed, whether that is a sentence, paragraph or a larger text. 
When calculating output probabilities over classes and words, we also condition them on this new document vector:
$$
classes = softmax(W_c \cdot hidden_t + W_{dc} \cdot doc)
$$
$$
output = softmax(W_o^{(c)} \cdot hidden_t + W_{do}^{(c)} \cdot doc)
$$

\noindent where $W_{dc}$ is the weight matrix between the document vector and class layer, and $W_{do}^{(c)}$ is the weight matrix between the document vector and output words in class $c$.

We construct the document vector by treating the values as parameters and optimising them during both training and testing using backpropagation.
At each time step, the system first performs a forward pass through the network and outputs probability distributions over classes and words. We then use the next word in the sequence to calculate the error derivatives in the output and backpropagate them back into the document vector. The update is not able to to affect the output at the current time step, but it will modify the document vector which will be used in the next time step. The same word that is used for modifying the document vector for the next time step is also available in the input layer of the next time step, therefore the system receives no additional knowledge as input.

We are interested in modelling individual sentences, therefore at the beginning of each sentence the document vector is reset to a specific starting state, which is optimised during training and shared between all sentences.
During testing, the values in the document vector are continuously modified depending on the error derivatives being backpropagated from the output layer, while all other parameters in the model stay constant.

When dealing with larger texts and domain-specific corpora, similar ideas of iterative learning can be applied to any language model. After processing a certain amount of data during testing, a new model could be trained using the previously seen testing examples as additional training data. Since this process adds more training data which is likely to be similar to upcoming testing examples, the system is likely to achieve a better performance.

However, when dealing with independent sentences, online learning becomes more difficult to apply. Each sentence contains very little additional data, and even if the language model is adjusted after every individual word, it only obtains evidence of previous words in the sentence, whereas these words are relatively unlikely to occur again in the same sentence. Therefore, instead of adjusting individual word representations, our approach learns a distributed document vector to represent the specific unit of text that is currently being processed. This vector is then used as additional evidence when calculating output probabilities.

\newcite{Le2014a} use a similar method for learning vector representations of documents and paragraphs. They construct a feedforward language model and include a paragraph vector as an additional vector in the input layer. The model parameters are trained on the training set, and when given unseen test data, the system optimises the paragraph vector according to the error signal. They use these vectors as input to a logistic regression classifier and achieve state-of-the-art performance on sentiment classification of movie reviews. However, they did not consider the effect of this model modification directly on the task of language modelling.

While the system of \newcite{Le2014a} uses a basic feedforward language model, we extend the idea to recurrent neural network language models, as they are currently used in state-of-the-art language modelling systems \cite{Chelba2014}. Attaching the document vector to the input layer is not preferable for RNNLM, as the error is only backpropagated into the input layer after several time steps. When this time step is reached and the network is unrolled to perform backpropagation through time, several words have already passed without receiving any additional information. Since our implementation performs the unrolling only at the end of each sentence, the updates would not have any effect. Therefore, we attach the document vector directly to the output layer, in parallel with the recurrent hidden component. Parameters in the document vector can then be updated at each time step, while the unrolling and backpropagation through time still happens at the end of the sentence.

\begin{table*}[t]
\centering
\fontsize{10.6pt}{14pt} \selectfont
\begin{tabular}{lccc|cc} \toprule
 & Train PPL & Dev PPL & Test PPL & +Parameters & +Operations \\ \midrule
Baseline M=100 & 92.65 & 103.56 & 102.51 & -- & -- \\ \midrule
M=120 & 88.60 & 98.78 & 97.79 & 666,960 & 7,400 \\
M=100, D=20 & \textbf{87.28} & \textbf{95.36} & \textbf{94.39} & \textbf{332,300} & \textbf{6,000} \\ \midrule
M=135 & 85.17 & 96.33 & 95.71 & 1,167,705 & 13,475 \\
M=100, D=35 & \textbf{80.11} & \textbf{91.05} & \textbf{90.29} & \textbf{581,525} & \textbf{10,500} \\ \bottomrule
\end{tabular}
\caption{Perplexity and additional parameters/operations for different language model configurations}
\label{tab:results}
\end{table*}

\section{Experiments}

We constructed a dataset from English Wikipedia to evaluate language modelling performance over individual sentences.
The text was tokenised, sentence split and lowercased. The sentences were shuffled, in order to minimise any transfer effects between consecutive sentences, and then split into training, development and test sets. The final sentences were sampled randomly, in order to obtain reasonable training times for the experiments. The dataset sizes are shown in Table \ref{tab:dataset_sizes}.

\begin{table}[h]
\centering
\fontsize{10.6pt}{14pt} \selectfont
\begin{tabular}{lccc} \toprule
& Train & Dev & Test \\ \midrule
Words & 9,990,782 & 237,037 & 4,208,847 \\
Sentences & 419,278 & 10,000 & 176,564 \\ \bottomrule
\end{tabular}
\caption{Dataset sizes}
\label{tab:dataset_sizes}
\end{table}

Model performance is measured using perplexity, therefore lower values indicate a model which is able to better predict the data.
Special tokens are used to mark the beginning and end of a sentence. The sentence end token is also included in the evaluation, whereas the sentence start token is only used as context in the input layer. Any words that occur less than 30 times in the training data were replaced by a special token for unknown words, leaving a vocabulary of 16,514 unique words.
General learning rate was set to $0.1$ and decreased during training, whereas the learning rate of the document vector was fixed at $0.1$ for both training and testing.

As the baseline, we use the regular RNNLM with 100-dimensional hidden layers and word vectors ($M = 100$). In the experiments we increase the capacity of the model and measure how that affects the perplexity on the datasets. First, we increase the value of M, allowing more information to be stored into word representations, while also increasing the number of hidden-hidden and hidden-output connections. As can be seen in Table \ref{tab:results}, this improves the overall performance of the model -- setting $M$ to $120$ and $135$ leads to progressively lower perplexity. 

Next, instead of increasing $M$, we add a $D$-dimensional document vector to the model and use this for online learning. When the same number of elements is added to $M$ or $D$, our results show consistently better performance when using the document vector. Increasing $M$ by 35 gives perplexity $95.71$, whereas using a 35-dimensional document vector gives perplexity $90.29$. We also performed the same experiment using only half of the training data, and the difference was even larger -- $105.50$ and $98.23$ correspondingly.

One reason why online learning during model deployment is not commonly used is because it is computationally expensive. Continuously retraining the model and adjusting parameters can be very time-consuming compared to a simple feedforward process through the network. However, extra computation is also needed when using a hidden vector of size $M$, as opposed to using a smaller value. When increasing the value of $M$ to $M+X$, the RNNLM will contain
$$
X \cdot C + 2 \cdot X \cdot V + 2 \cdot X \cdot M + X^2
$$

\noindent additional parameters and needs to perform
$$
2 \cdot X \cdot M + X^2 + X \cdot C + X \cdot E[O]
$$

\noindent additional operations at each time step.\footnote{We only count the matrix multiplication operations, as they take the majority of the time in a neural network language model.} $C$ is the number of classes, $V$ is vocabulary size, and $E[O]$ is the expected number of words that need to be processed in the output layer during one step.

The corresponding number of additional parameters in a RNNLM model using a $D$-dimensional document vector for online learning is
$$
D + D \cdot V + D \cdot C
$$

\noindent and additional operations
$$
2 \cdot D \cdot E[O] + 2 \cdot D \cdot C
$$

\noindent which includes the error backpropagation at each time step. For our experiments $V = 16,514$, $C = 100$ and $E[O] \approx 50$. Table \ref{tab:results} contains the additional values for the experiments, showing that replacing some hidden vector parameters with the actively learned document vector leads to fewer total parameters and fewer operations, along with lower perplexity.

\begin{table*}[tbh]
\centering
\begin{tabular}{p{16cm}} \toprule
Both Hufnagel and Marston also joined the long-standing technical death metal band Gorguts.
\vspace{-2mm}
\begin{enumerate}
\item The band eventually went on to become the post-hardcore band Adair. 
\vspace{-2.4mm}
\item The band members originally came from different death metal bands, bonding over a common interest in d-beat. 
\vspace{-2.4mm}
\item The proceeds went towards a home studio, which enabled him to concentrate on his solo output and songs that were to become his debut mini-album "Feeding The Wolves". 
\vspace{-2.4mm}
\end{enumerate}
\\ \midrule
The Chiefs reclaimed the title on September 29, 2014 in a Monday Night Football game against the New England Patriots, hitting 142.2 decibels.
\vspace{-2mm}
\begin{enumerate}
\item He played in twenty-four regular season games for the Colts, all off the bench.
\vspace{-2.4mm}
\item In May 2009 the Warriors announced they had re-signed him until the end of the 2011 season.
\vspace{-2.4mm}
\item The team played inconsistently throughout the campaign from the outset, losing the opening two matches before winning four consecutive games during September 1927.
\vspace{-2.4mm}
\end{enumerate}
\\ \midrule
He was educated at Llandovery College and Jesus College, Oxford, where he obtained an M.A. degree.
\vspace{-2mm}
\begin{enumerate}
\item He studied at the Orthodox High School, then at the Faculty of Mathematics.
\vspace{-2.4mm}
\item Kaigama studied for the priesthood at St. Augustine's Seminary in Jos with further study in theology in Rome.
\vspace{-2.4mm}
\item Under his stewardship, Zahira College became one of the leading schools in the country.
\vspace{-2.4mm}
\end{enumerate}
\\
\bottomrule
\end{tabular}
\caption {Examples of using the document vectors to find similar sentences in the development data.} 
\label{tab:examples}
\end{table*}

\begin{figure}[h]
    \includegraphics[width=\linewidth]{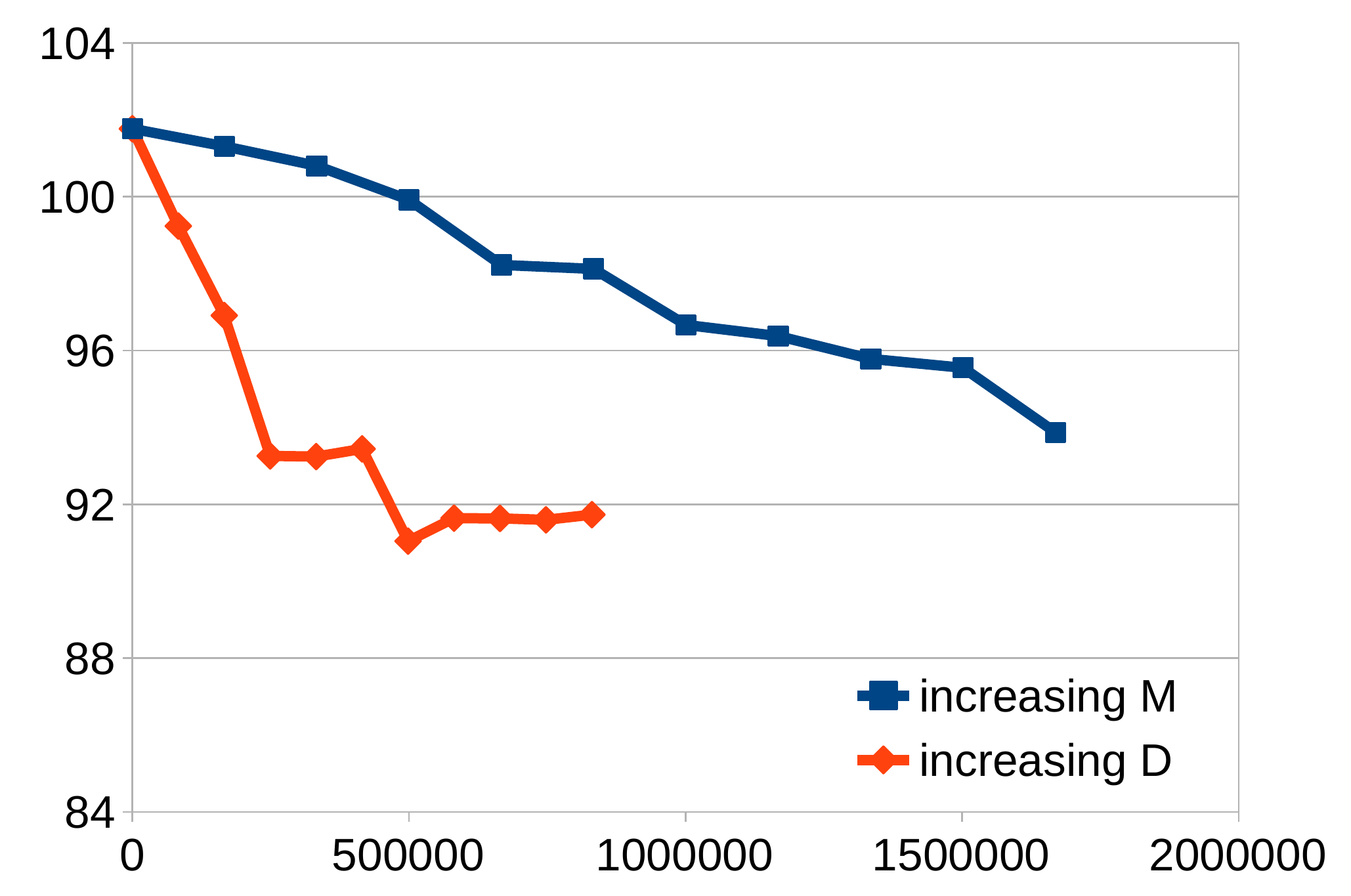}
	\caption{Perplexity as a function of additional parameters when increasing either $M$ or $D$. The x-axis shows the number of additional parameters in the model, with respect to the baseline of $M=100$, $D=0$. The y-axis shows the perplexity on the test set.}
	\label{fig:param_vs_ppl}
\end{figure}

Figure \ref{fig:param_vs_ppl} presents the relationship between perplexity and the number of additional parameters, when increasing either $M$ or $D$. The results are averaged over 10 runs with different random initialisations. As can be seen, using a small document vector lowers the perplexity with fewer parameters, compared to simply increasing the main components of the network. The graph of perplexity with respect to additional operations in the model also has a very similar shape.

In order to further explore the relationship between $D$ and $M$, we trained a number of smaller models with different values, under the constraint $D + M = 100$. To reduce computation time, only half of the training data was used in these experiments. The lowest perplexity was achieved in the region of $D = 23$ and $M = 77$, and making the document vectors much smaller or larger led to a decrease in performance. 
This indicates that including the document vector does help increase model accuracy, but as it contains no information about the training data, this vector should be small compared to the main model.

Intuitively, this approach works by having the document vector capture the unique aspects of each sentence. While the general RNNLM is a smooth static representation of the entire training data, the document vector is optimised to represent how each sentence differs from the main language model. Therefore we performed a qualitative evaluation and found that the learned sentence vectors were also very good predictors of semantic similarity.
The RNN language model was trained on the training set, and then used to process the development set. The last state of the document vector of each sentence was used to calculate cosine similarity.
Table \ref{tab:examples} contains randomly sampled sentences from the development set, together with corresponding development sentences that have the highest similarity (excluding the original sentence). Even though there is almost no word overlap, the retrieved sentences are semantically very similar.

\section{Conclusion}

We have described a possible extension of RNNLM which uses continuous online learning. The model includes a separate vector to represent the unit of text, such as a sentence, being currently processed. The vector starts in a default state and is continuously updated using backpropagation, leading to a more informative representation. 
The modified language model achieves lower perplexity with a more optimal use of parameters.

The idea of continuous training and adaptation is natural and also established in biological learning processes, yet it is not widely used due to computational complexity. 
Our experiments indicate that by including this active learning component in the neural network model, the system is able to achieve higher accuracy, while also decreasing the parameters needed to store the model and decreasing the computation required.

\bibliographystyle{acl}
\bibliography{references}

\end{document}